\documentclass[conference]{IEEEtran}
\usepackage{cite}
\usepackage{amsmath,amssymb,amsfonts}
\usepackage{algorithmic}
\usepackage{graphicx}
\usepackage{textcomp}
\usepackage{xcolor}
\usepackage{booktabs}
\usepackage[font=small,skip=5pt]{caption} %Get rid of this to adhere to IEEE 
\def\BibTeX{{\rm B\kern-.05em{\sc i\kern-.025em b}\kern-.08em
    T\kern-.1667em\lower.7ex\hbox{E}\kern-.125emX}}
\begin{document}

\title{Towards Goal, Feasibility, and Diversity-Oriented Deep Generative Models in Design\\
}
\author{\IEEEauthorblockN{Lyle Regenwetter}
\IEEEauthorblockA{\textit{Deparment of Mechanical Engineering} \\
\textit{Massachussetts Institute of Technology}\\
regenwet@mit.edu}
\and
\IEEEauthorblockN{Faez Ahmed}
\IEEEauthorblockA{\textit{Deparment of Mechanical Engineering} \\
\textit{Massachussetts Institute of Technology}\\
faez@mit.edu}
}

% \author{\IEEEauthorblockN{Lyle Regenwetter\IEEEauthorrefmark{1},
% Faez Ahmed\IEEEauthorrefmark{2}}
% \IEEEauthorblockA{Deparment of Mechanical Engineering\\
% Massachussetts Institute of Technology\\
% Email: \IEEEauthorrefmark{1}regenwet@mit.edu,
% \IEEEauthorrefmark{2}faez@mit.edu}}
% \maketitle

\maketitle

%%%%%%%%%%%%%%%%%%%%%%%%%%%%%%%%%%%%%%%%%%%%%%%%%%%%%%%%%%%%%%%%%%%%%%
\section{abstract}
{\it 
Deep Generative Machine Learning Models (DGMs) have been growing in popularity across the design community thanks to their ability to learn and mimic complex data distributions. DGMs are conventionally trained to minimize statistical divergence between the distribution over generated data and distribution over the dataset on which they are trained. While sufficient for the task of generating ``realistic'' fake data, this objective is typically insufficient for design synthesis tasks. Instead, design problems typically call for adherence to design requirements, such as performance targets and constraints. Advancing DGMs in engineering design requires new training objectives which promote engineering design objectives. In this paper, we present the first Deep Generative Model that simultaneously optimizes for performance, feasibility, diversity, and target achievement. We benchmark performance of the proposed method against several Deep Generative Models over eight evaluation metrics that focus on feasibility, diversity, and satisfaction of design performance targets. Methods are tested on a challenging multi-objective bicycle frame design problem with skewed, multimodal data of different datatypes. The proposed framework was found to outperform all Deep Generative Models in six of eight metrics. 
}

\setlength{\belowdisplayskip}{5pt} \setlength{\belowdisplayshortskip}{5pt}
\setlength{\abovedisplayskip}{5pt} \setlength{\abovedisplayshortskip}{5pt}

%%%%%%%%%%%%%%%%%%%%%%%%%%%%%%%%%%%%%%%%%%%%%%%%%%%%%%%%%%%%%%%%%%%%%%

\section{Introduction}
Automatically creating innovative designs that outperform all existing solutions and meet complex real-world engineering constraints is the holy grail of data-driven engineering design. This is an incredibly demanding task and current design automation tools remain insufficient for full autonomy in product design. Recently, Deep Generative Models (DGMs) have emerged as a viable means to bridge the gap to this overarching design automation goal. DGMs, such as Generative Adversarial Networks and Variational Autoencoders, are typically trained to minimize the statistical divergence (or maximize similarity) between distributions of generated samples and the underlying data distribution. In engineering design, design objectives and constraints make statistical similarity metrics insufficient and sometimes inappropriate. Despite this, an overwhelming majority of research in engineering design continues to optimize and evaluate methods using statistical similarity. We believe the continuation of this practice is rooted in two central challenges. Firstly, appropriate metrics to evaluate DGMs on engineering objectives such as design performance, feasibility, and achievement of design requirements are poorly established. Secondly, researchers lack effective methods to build these auxiliary objectives into training procedures and instead fall back upon the established structural similarity as the central training mechanism. This paper presents our proposed solution to the aforementioned challenges. Our key contributions are summarized as follows:
\begin{enumerate}
    \item We introduce the Design Target Achievement Index (DTAI), a differentiable loss function which allows Deep Generative Models to prioritize, meet, and exceed multi-objective performance targets specified by a designer. 
    \item We introduce the first Deep Generative Model that simultaneously optimizes for design performance, diversity, feasibility, and target satisfaction. We demonstrate that our specialized loss function yields significant performance improvements, such as increasing the average proportion of design targets met by 45\% and the proportion of feasible designs by 30\% versus state-of-the-art tabular generation methods. 
    \item We demonstrate that the proposed framework outperforms GANs, Conditional Tabular GANs, and Multi-Objective Performance-aware Diverse GANs in six evaluation metrics tested.
\end{enumerate}
% In the following sections, we briefly review Deep Generative Models in engineering design, introduce and test our framework, and analyze results. 

\section{Review of Deep Generative Models in Engineering Design}
In a recent review of Deep Generative Models (DGMs), Regenwetter et al.~\cite{regenwetter2021deep} discuss the application of DGMs across engineering design fields and analyze key limitations in the current state-of-the-art in DGM methodology. For a more detailed review and discussion, we refer the reader to ~\cite{regenwetter2021deep}. The authors suggest that successfully addressing several key challenges will be essential in the continued development of DGMs for engineering design. Four of these challenges are design performance, design novelty, more robust design representation methods, and targeted inverse design.

Design performance (quality) may be comprised of many diverse (and sometimes competing) objectives, such as weight, cost, aerodynamics, etc. Several approaches to incorporate design performance into DGMs have been proposed: 1) Building performance-estimating objectives into the training function ~\cite{chen2021padgan, chen2021mopadgan}; 2) Iteratively training the DGM on datasets augmented with high-performing generated designs~\cite{oh2019deep, shu20203d}; 3) Optimizing latent vectors using a fitting surrogate model~\cite{rawat2019application, guo2018indirect}. Despite this, existing DGMs in engineering design lack any mechanism to condition design generation toward a specific set of designer-specified performance targets. Our proposed framework uses an auxiliary loss (approach \#1) that considers performance, target achievement, diversity, and feasibility of generated designs in a single aggregate score.

\section{Methodology}
\begin{figure}
    \centering
    \includegraphics[width=\linewidth]{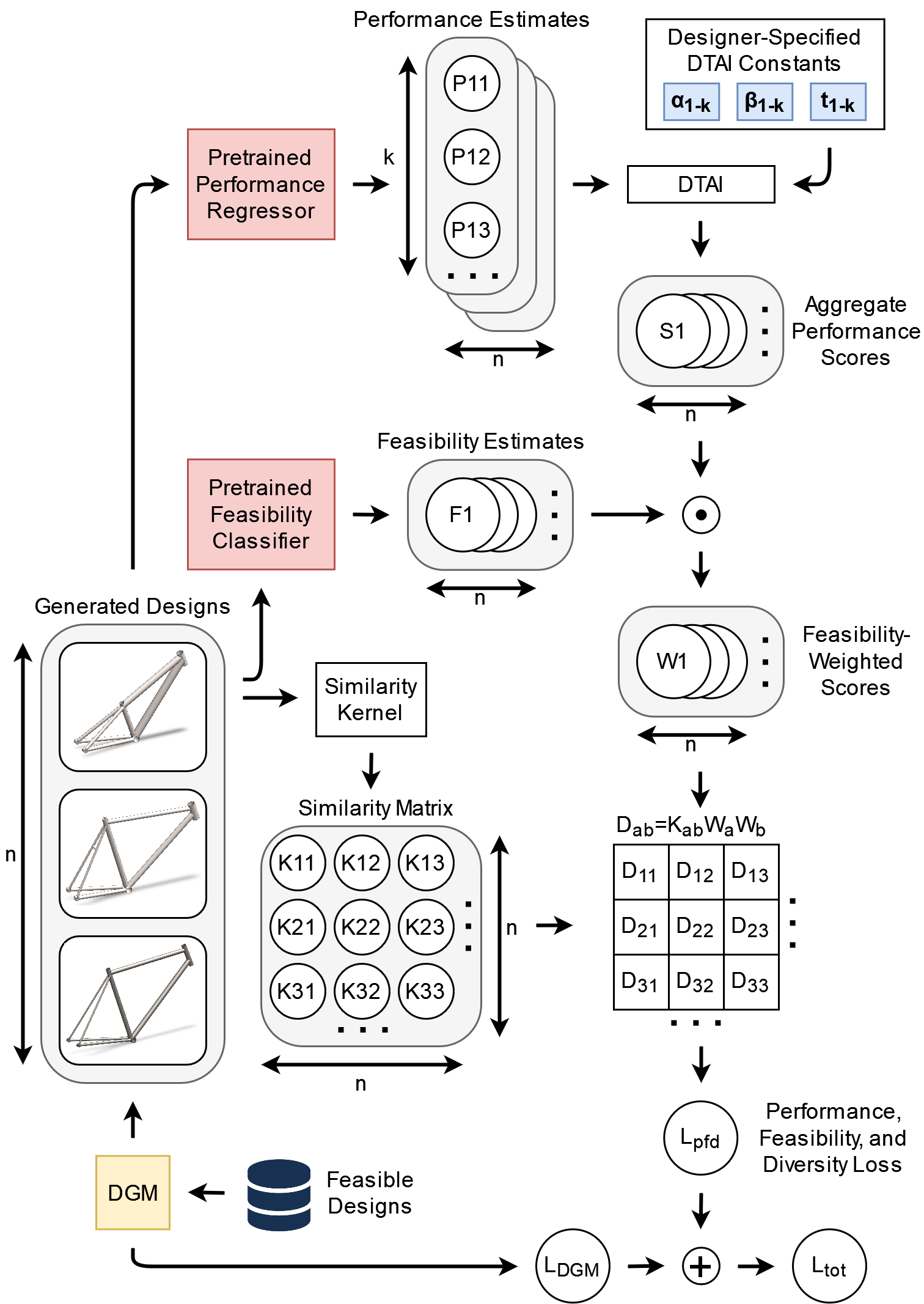}
    \caption{Proposed Performance, Feasibility, and Diversity-Aware DGM}
    \label{fig:flow}
\end{figure}

To construct a loss function that considers performance, target achievement, diversity, and feasibility, we first propose an aggregate performance score that considers a design's adherence to designer-specified design performance targets. We then augment this score with a feasibility likelihood for a feasibility-weighted performance score. Finally, we scale these feasibility-weighted performance scores with diversity scores and compute a final loss using a Determinantal Point Process, as proposed in ~\cite{chen2021padgan}. This loss is appended to the loss function of a Deep Generative Model. In our testing, we use a Generative Adversarial Network. The overall pipeline is shown in Figure~\ref{fig:flow}. Below, we discuss how the various scores are calculated. 
\subsection{Aggregate Performance Score} \label{sec:aggregate}
\begin{figure}
    \centering
    \includegraphics[width=0.5\textwidth]{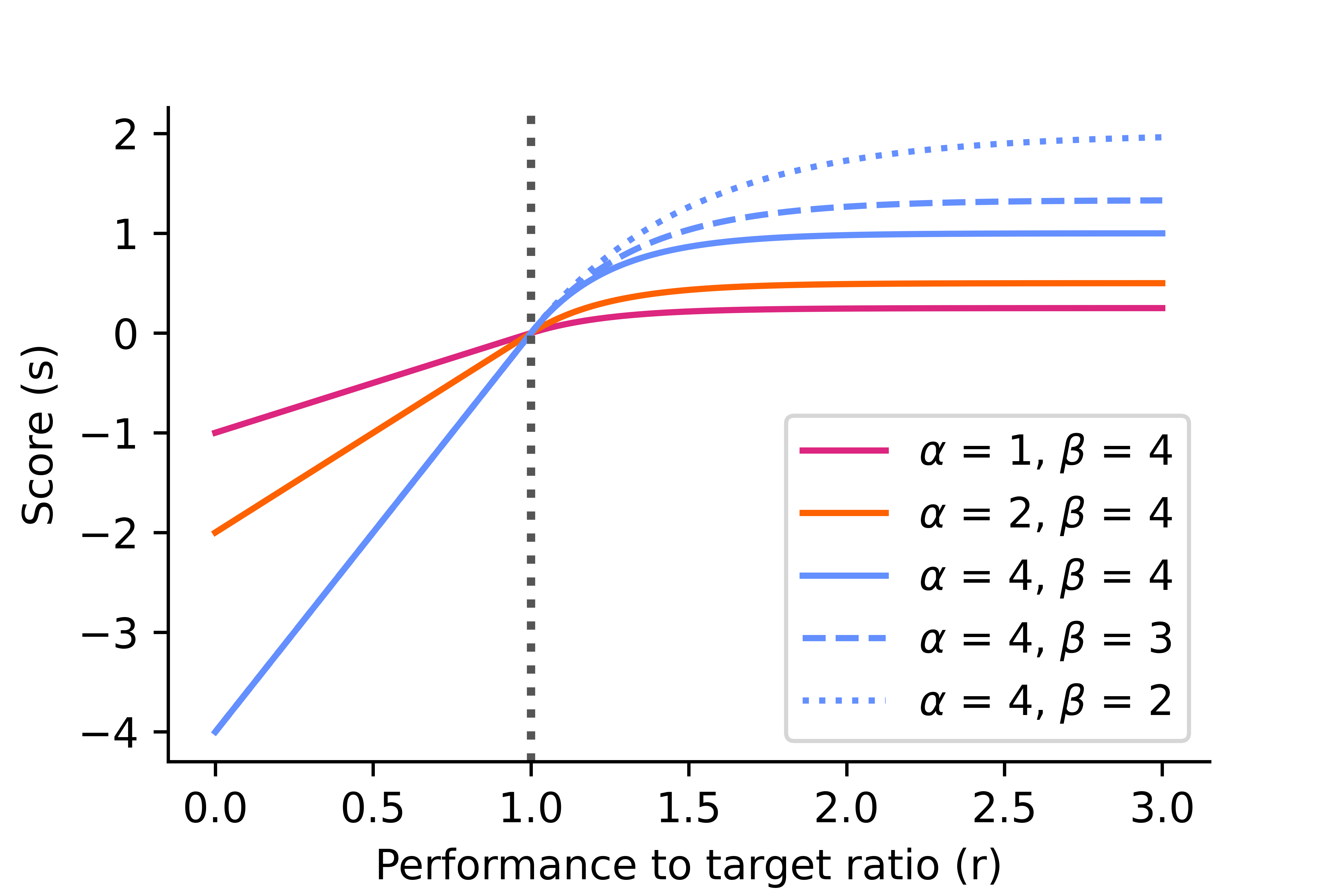}
    \caption{The $\alpha$ parameter adjusts the steepness of the Design Target Achievement Index (DTAI) curve, while the $\beta$ parameter adjusts how steeply it falls off after the target is exceeded.}
    \label{fig:abmod}
\end{figure}

To quantify how well a design adheres to designer-specified performance targets, we propose a novel training loss called the Design Target Achievement Index (DTAI) in~\cite{regenwetter2022design}. We provide an overview of the loss here, but refer readers to the original paper for more details. Consider a design, $i$, and its performance $p_{i,k}$ with respect to a particular performance target, $t_k$, for $k\in\{1 ... T\}$ where $T$ is the number of design objectives (i.e. weight, safety factor, etc.). In any given objective, $k$, we desire our design's performance to exceed the performance target: $p_{i,k}\geq t_{k}$. When our design's performance exceeds the target, we have $r_{i,k}=\frac{p_{i,k}}{t_{k}}\geq1$. We propose the following piecewise scoring function to compute an individual target achievement score, $s_{i,k}$ in terms of $r_{i,k}$. This function is parameterized by two tuning factors, $\alpha_k$ and $\beta_k$, which adjust the steepness of the curve (importance of the objective) and the decay once the target is exceeded (how valuable improvement beyond the target is for this objective). Figure~\ref{fig:abmod} shows the effect of adjusting $\alpha_k$ and $\beta_k$. 
\begin{equation}
    s_{i,k}=\begin{cases} 
          \alpha_k(1-r_{i,k}) & r_{i,k}\leq 1 \\
          \frac{\alpha_k}{\beta_k}(1-e^{\beta_k(1-r_{i,k})}) & r_{i,k}> 1 
    \end{cases}
\end{equation}

The individual scores $s_{i,k}$ for each objective are summed into an aggregate score, then scaled by the minimum and maximum values for the sum, $s_{min}$ and $s_{max}$, which can be easily derived. This final result is our proposed Design Target Achievement Index, $s_{DTAI}$.
\begin{equation}
    s_{DTAI, i}=\frac{(\sum_{k=1}^n s_{i,k}) -s_{min}}{s_{max}-s_{min}}
\end{equation}
\begin{equation}
    s_{min}=-\sum_{k=1}^T \alpha_k, \,\,\, s_{max}=\sum_{k=1}^T \frac{\alpha_k}{\beta_k}
\end{equation}

DTAI has many desirable properties: DTAI's derivative with respect to an individual objective is large when a design is currently underperforming the objective's performance target but exponentially decays as the design outperforms the target. DTAI is differentiable across the entire space of possible performance and constraint values and its derivative is continuous. Both DTAI and its derivative are bounded. DTAI's computational cost scales linearly with the number of design objectives. Finally, DTAI's $\alpha$ and $\beta$ parameters enable easy and precise customization by the designer. 

Since DTAI is used as a trianing loss, performance estimates must be differentiable. In the example application, we pretrain a supervised regressor to estimate performance for generated designs.

\subsection{Feasibility-Weighted Performance Scores}
DGMs in engineering design often struggle with physical infeasibility of generated designs (conflicting geometry, negative distances, etc.). Just as we train a surrogate model to predict performance of generated designs, we can also predict their feasibility using a pretrained supervised classifier. Given predicted feasibility likelihood for a generated design, we multiply its aggregate performance score with its feasibility likelihood for a feasibility-weighted performance score. 
\subsection{Diversity, Feasibility, and Performance Loss}
Chen \& Ahmed~\cite{chen2021padgan} propose a method to incorporate performance and diversity into DGM training using a weighted similarity matrix calculated with a similarity kernel and a Determinantal Point Process (DPP). We borrow the proposed approach, calculating a similarity matrix, which we scale using feasibility-weighted performance scores and a hyperparameter that modulates the imporatnce of diversity versus performance. The final diversity, feasibility, and performance loss is calculated using a DPP and appended to the DGM's standard training loss.

\section{Experimental Results} 
\subsection{Dataset}
For this study, we select the recently-introduced FRAMED dataset~\cite{regenwetter2022framed}, which consists of 4500 community-designed bicycle frame models~\cite{BIKED}, parameterized over 37 design variables (tube dimensions, frame material, etc). The FRAMED dataset provides a set of ten structural performance measures for each frame (weight, deflections, etc.). For testing, we select a demanding set of performance targets equal to the $75^{th}$ percentile performance scores from the dataset in each objective. 

\subsection{Benchmark Models}
We benchmark our proposed framework against several baselines. The simplest of these are random sampling from the dataset and random interpolation between dataset models. We also benchmark a vanilla GAN and a Conditional Tabular GAN (CTGAN) proposed in~\cite{xu2019modeling}, which do not consider performance, diversity, feasibility or target achievement. Finally, we benchmark a Multi-Objective Performance-Aware Diverse GAN (MO-PaDGAN), as proposed in~\cite{chen2021mopadgan}, which considers performance and diversity, but not feasibility or target achievement. When introduced, MO-PaDGAN improved multi-objective performance for aerodynamic synthesis by 186\% compared to a standard GAN, and thus constitutes a strong model for comparison.

\subsection{Results}
\begin{figure}[!htb]
    \centering
    \includegraphics[width=\linewidth]{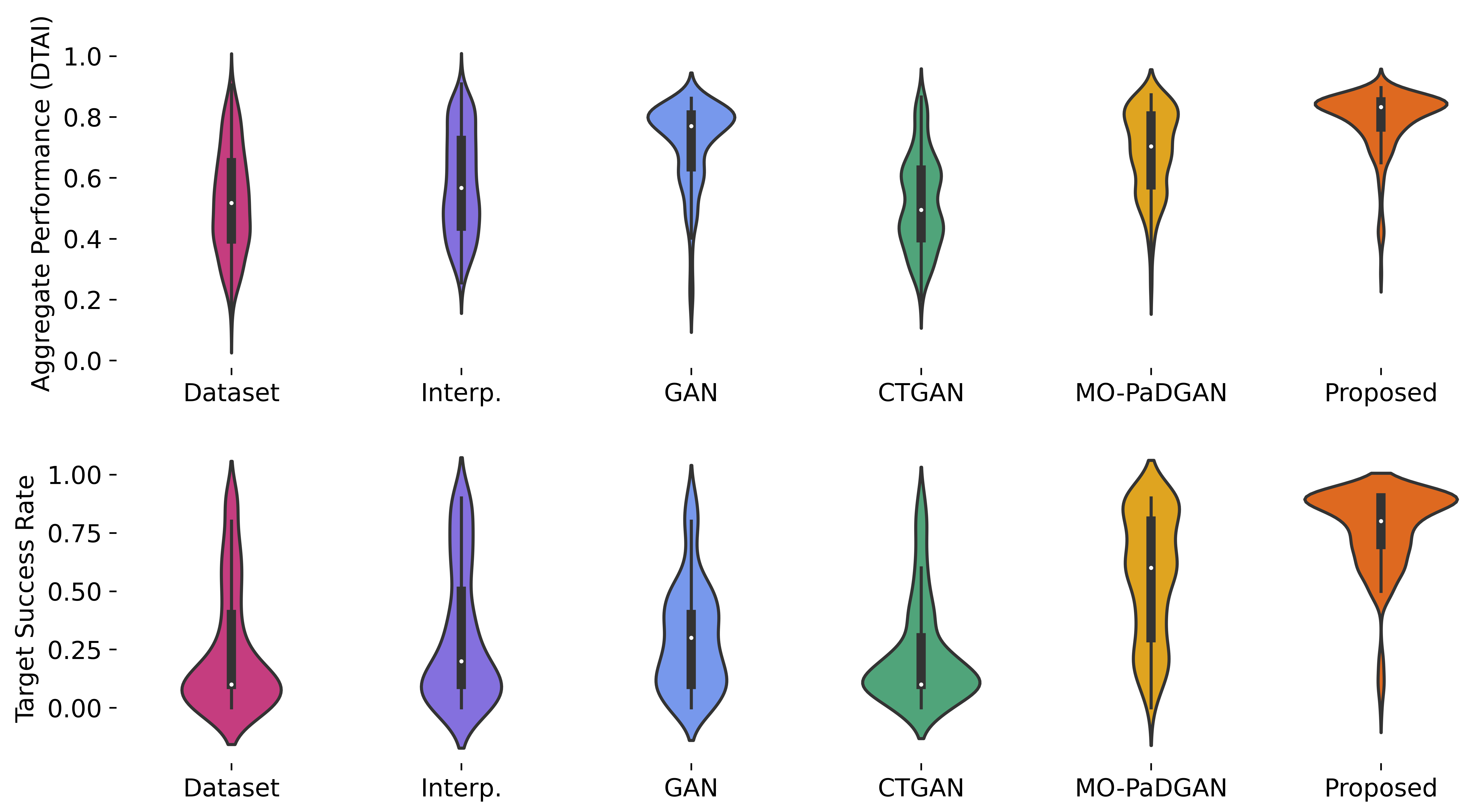}
    \caption{Violin plots demonstrate that the distribution of designs synthesized the proposed method tend toward significantly higher performance and target success rate compared to distributions of designs synthesized by baseline models. 
    }
    \label{fig:violins}
\end{figure}

Table~\ref{tab:results} shows the results of our testing. All in all, the proposed framework achieves superior performance in all evaluation metrics except diversity over all other Deep Generative Models tested. While diversity scores are modest, diversity can be prioritized using the diversity modulation hyperparameter, if desired. The framework boasts a $95.9\%$ feasibility rate among generated designs over average the average $72.7\%$ across the baseline DGMs. Furthermore, it attains an average of $69.8\%$ of design targets as compared to an average of $33.8\%$ across baseline DGMs. Figure~\ref{fig:violins} presents violin plots of the distribution of Design Target Achievement Index and Target Success Rate over the space of generated designs for baseline methods. A continuous distribution is approximated over the 250 designs generated by each method using a Kernel Density Estimate (KDE). The proposed framework is shown to consistently outperform competing methods across distributions of generated samples in target achievement and aggregate performance. 

\begin{table*}[!htb]
\centering
\caption{Deep Generative Models scored on the six proposed evaluation metrics. Models from left to right: Randomly sampled design subsets from the FRAMED dataset (Dataset), Random Interpolation between FRAMED designs (Interpolation), Vanilla GAN (GAN), Conditional Tabular GAN (CTGAN), Proposed PaDGAN with DTAI and auxiliary Geometric Feasibility Classifier (Proposed)}
\label{tab:results}
\resizebox{0.95\textwidth}{!}{%
\begin{tabular}{lcccccc}
\toprule
\textbf{} & Dataset & Interpolation & \begin{tabular}[c]{@{}c@{}}GAN\\ (2014)\end{tabular} & \begin{tabular}[c]{@{}c@{}}CTGAN\\ (2019)\end{tabular} & \begin{tabular}[c]{@{}c@{}}MO-PaDGAN\\ (2020)\end{tabular} & {\color[HTML]{9E0000} \begin{tabular}[c]{@{}c@{}}Proposed\\ (2022)\end{tabular}} \\ \midrule
Mean Target Success Rate (TSR) (\%) & 24 & {\color[HTML]{BF8F00} 31.3} & {\color[HTML]{C65911} 28.6} & {\color[HTML]{C00000} 24.4} & {\color[HTML]{548235} 48.4} & {\color[HTML]{7CB953} 69.8} \\ 
Feasibility Rate (GFR) (\%) & 100 & {\color[HTML]{7CB953} 100} & {\color[HTML]{C00000} 65.2} & {\color[HTML]{C65911} 65.7} & {\color[HTML]{BF8F00} 87.1} & {\color[HTML]{548235} 95.9} \\ 
Mean Design Target Achievement Index   (DTAI) & 0.53 & {\color[HTML]{C65911} 0.58} & {\color[HTML]{548235} 0.71} & {\color[HTML]{C00000} 0.52} & {\color[HTML]{BF8F00} 0.69} & {\color[HTML]{7CB953} 0.79} \\ 
Mean Minimum Target Ratio (MTR) & 0.32 & {\color[HTML]{C65911} 0.37} & {\color[HTML]{BF8F00} 0.41} & {\color[HTML]{C00000} 0.32} & {\color[HTML]{548235} 0.42} & {\color[HTML]{7CB953} 0.43} \\ 
Hypervolume (HV) *10\textasciicircum{}-7 & 4.81 & {\color[HTML]{7CB953} 4} & {\color[HTML]{C65911} 3.04} & {\color[HTML]{BF8F00} 3.54} & {\color[HTML]{C00000} 2.95} & {\color[HTML]{548235} 3.82} \\ 
Mean Design Novelty (DN) & 0 & {\color[HTML]{C00000} 7.17} & {\color[HTML]{BF8F00} 8.3} & {\color[HTML]{C65911} 8.02} & {\color[HTML]{548235} 9.94} & {\color[HTML]{7CB953} 9.96} \\ 
Mean Design Space Diversity & 8.6 & {\color[HTML]{548235} 13.28} & {\color[HTML]{BF8F00} 10.08} & {\color[HTML]{7CB953} 14.12} & {\color[HTML]{C00000} 7.11} & {\color[HTML]{C65911} 9.48} \\
Mean Performance Space Diversity (PSD) & 3.8 & {\color[HTML]{548235} 3.4} & {\color[HTML]{BF8F00} 2.71} & {\color[HTML]{7CB953} 3.95} & {\color[HTML]{C00000} 2.62} & {\color[HTML]{BF8F00} 2.71} \\

\bottomrule
\end{tabular}%
}
\end{table*}

\subsection{Ablation Study} 
The ablation study shown in Table~\ref{tab:ablation} contrasts the proposed PaDGAN with DTAI and auxiliary classifier against variants without the classifier (-CLF) and using MO-PaDGAN's random objective weighting instead of DTAI (-DTAI). We note that the configuration without the classifier and using MO-PaDGAN's random objective weighting is equivalent to the previous state-of-the-art MO-PaDGAN (-DTAI, -CLF). We outperform Mo-PaDGAN in every metric tested, with significant improvements in feasibility (95.9\% from 87.1\%), DTAI (0.79 from 0.69), Target Success Rate (TSR) (69.8\% from 48.4\%), and Hypervolume ($3.82E-7$ from $2.95E-7$).

\begin{table*}[!htb]
\centering
\caption{Ablation Study Contrasting Proposed PaDGAN with DTAI training loss and auxiliary classifier against PaDGAN without DTAI (-DTAI), without the auxiliary classifier (-CLF), and MO-PaDGAN (-DTAI, -CLF). }
\label{tab:ablation}
\resizebox{0.75\textwidth}{!}{%
\begin{tabular}{lcccc}
\toprule
\textbf{} & Proposed & -DTAI & -CLF & -DTAI, -CLF \\ \midrule
Mean Target Success Rate (\%) & {\color[HTML]{098D00} 69.8} & {\color[HTML]{8DA500} 67.7} & {\color[HTML]{A57B00} 51.4} & {\color[HTML]{A53300} 48.4} \\ 
Feasibility Rate  (\%) & {\color[HTML]{098D00} 95.9} & {\color[HTML]{A53300} 82.5} & {\color[HTML]{A57B00} 83.2} & {\color[HTML]{8DA500} 87.1} \\ 
Mean Minimum Target Ratio & {\color[HTML]{A57B00} 0.43} & {\color[HTML]{098D00} 0.51} & {\color[HTML]{8DA500} 0.46} & {\color[HTML]{A53300} 0.42} \\
Mean Design Target Achievement Index (DTAI) & {\color[HTML]{8DA500} 0.79} & {\color[HTML]{098D00} 0.80} & {\color[HTML]{A57B00} 0.72} & {\color[HTML]{A53300} 0.69} \\ 
Hypervolume *10\textasciicircum{}{(-7)} & {\color[HTML]{098D00} 3.82} & {\color[HTML]{098D00} 3.82} & {\color[HTML]{A57B00} 3.40} & {\color[HTML]{A53300} 2.95} \\ 
Mean Design Novelty & {\color[HTML]{8DA500} 9.96} & {\color[HTML]{A53300} 9.90} & {\color[HTML]{098D00} 10.52} & {\color[HTML]{A57B00} 9.94} \\ 
Mean Design Space Diversity & {\color[HTML]{098D00} 9.48} & {\color[HTML]{A57B00} 6.21} & {\color[HTML]{A53300} 5.80} & {\color[HTML]{8DA500} 7.11} \\ 
Mean Performance Space Diversity & {\color[HTML]{098D00} 2.71} & {\color[HTML]{A53300} 2.24} & {\color[HTML]{A57B00} 2.44} & {\color[HTML]{8DA500} 2.62} \\ 
\bottomrule
\end{tabular}%
}
\end{table*}

\section{Conclusion}

We introduce a novel differentiable scoring metric called Design Target Achievement Index (DTAI) which allows Deep Generative Models to prioritize, meet, and exceed multi-objective performance targets. We augment a Performance-Augmented Diverse GAN with our DTAI objective and demonstrate significantly improved performance in design generation. We then further augment this PaDGAN with an auxiliary classifier to encourage the generation of feasible results. To benchmark our method, we evaluate a variety of Deep Generative Models, including the Multi-Objective PaDGAN, and specialized tabular generation algorithm, CTGAN. Methods are tested on a challenging bicycle frame design problem with 10 performance objectives. The proposed PaDGAN with DTAI loss and auxiliary classifier outperforms baseline DGMs in every evaluation metric and further outperforms other PaDGAN variants in ablation studies. All in all, this work establishes a novel Deep Generative Framework that actively optimizes performance, diversity, feasibility, and target satisfaction to establish a new state-of-the-art in design generation using Deep Generative Models. 

\section{Acknowledgments}
We would like to thank Amin Heyrani Nobari for creating the Tensorflow 2.x version of PaDGAN which we modified to generate our results. We also acknowledge MathWorks for supporting this research.

\bibliographystyle{IEEEtran}
\bibliography{bibliography}

\newpage

\end{document}